\pdfoutput=1

\documentclass[11pt]{article}

\usepackage[final]{acl}

\usepackage{times}
\usepackage{latexsym}

\usepackage[T1]{fontenc}

\usepackage[utf8]{inputenc}

\usepackage{microtype}

\usepackage{inconsolata}

\usepackage{graphicx}

\usepackage{array} 
\usepackage{multirow}  
\usepackage{amsmath}  
\usepackage{booktabs}  

\title{Advancing Semantic Textual Similarity Modeling: A Regression Framework with Translated ReLU and Smooth K2 Loss}

\author{Bowen Zhang \and Chunping Li \\
  School of Software, Tsinghua University \\
  \texttt{zbw23@mails.tsinghua.edu.cn}, \texttt{cli@tsinghua.edu.cn} \\}

\begin{document}
\maketitle
\begin{abstract}
Since the introduction of BERT and RoBERTa, research on Semantic Textual Similarity (STS) has made groundbreaking progress. Particularly, the adoption of contrastive learning has substantially elevated state-of-the-art performance across various STS benchmarks. However, contrastive learning categorizes text pairs as either semantically similar or dissimilar, failing to leverage fine-grained annotated information and necessitating large batch sizes to prevent model collapse. These constraints pose challenges for researchers engaged in STS tasks that involve nuanced similarity levels or those with limited computational resources, compelling them to explore alternatives like Sentence-BERT. Despite its efficiency, Sentence-BERT tackles STS tasks from a classification perspective, overlooking the progressive nature of semantic relationships, which results in suboptimal performance. To bridge this gap, this paper presents an innovative regression framework and proposes two simple yet effective loss functions: Translated ReLU and Smooth K2 Loss. Experimental results demonstrate that our method achieves convincing performance across seven established STS benchmarks and offers the potential for further optimization of contrastive learning pre-trained models. \footnote{Our code and checkpoints are available at \url{https://github.com/ZBWpro/STS-Regression}.}
\end{abstract}

\section{Introduction}

Semantic Textual Similarity (STS) constitutes a fundamental task in natural language processing, wielding significant influence across a multitude of applications, including text clustering, information retrieval, and recommendation systems. Despite the remarkable precision obtained by interactive architectures within these tasks, their inability to support offline computation limits their viability in large-scale text analysis scenarios. In response, the seminal work of Sentence-BERT \citep{Sentence-BERT-EMNLP-2019} introduces a dual-tower architecture to encode the sentences within a pair separately, thereby facilitating the derivation of independent embeddings. This approach showcases superior efficacy and has rapidly gained widespread acceptance, now serving as a cornerstone for various downstream tasks. Consequently, further improvements to Sentence-BERT hold significant research interest and practical value.

Nevertheless, the advent of contrastive learning methods, exemplified by SimCSE \citep{SimCSE-EMNLP-2021}, has led to more pronounced enhancements on renowned English STS benchmarks, such as STS12-16 \citep{STS12, STS13, STS14, STS15, STS16}, STS-B \citep{STS-B}, and SICK-R \citep{SICK-R}. This has shifted the research focus in recent years towards integrating contrastive learning techniques with pre-trained language models (PLMs) like BERT \citep{BERT-NAACL-2019} and RoBERTa \citep{RoBERTa-2019}. An intuitive comparison is that, when employing the NLI dataset \citep{snli-2015-EMNLP, mnli-2018-NAACL} as a training corpus, SimCSE-RoBERTa$_\text{base}$ attains an average Spearman's correlation score of 82.52 across these STS tasks, hugely surpassing the 74.21 achieved by Sentence-RoBERTa$_\text{base}$.

Such discernible performance disparity has inadvertently overshadowed the advantages of Sentence-BERT, especially in terms of data utilization efficiency and computational resource demands. Contrastive learning, by its self-supervised nature, predominantly recognizes text pairs as either similar or dissimilar. This binary categorization restricts contrastive learning methods to training on triplet-form data composed of an anchor sentence, a positive instance, and a hard negative instance in supervised settings \citep{SimCSE-EMNLP-2021}. Many practical scenarios, however, tend to provide more finely grained labeled data (e.g., highly relevant, moderately relevant, relevant, and irrelevant) \citep{RankCSE-ACL-2023}, where contrastive learning approaches can usually only exploit text pairs whose similarity indicators are at the endpoints.

Furthermore, since contrastive learning enhances model discriminability by treating other samples within the same batch as negative instances, it requires large batch sizes, thereby consuming substantial computational resources. For example, SimCSE's supervised learning settings include a batch size of 512 and 3 epochs. To accommodate this configuration on consumer-grade GPUs, SimCSE limits the maximum input length to 32 tokens \citep{SimCSE-EMNLP-2021}. In contrast, Sentence-BERT and our proposed methodology necessitate a mere batch size of 16 and 1 epoch to reach convergence. Additionally, our default maximum input length is 256, significantly longer than SimCSE's.

The aforementioned drawbacks highlight the difficulty in completely replacing Sentence-BERT with contrastive learning methods. Hence, some cutting-edge works \citep{AutoCoT-ICLR-2023} continue to rely on Sentence-BERT for sentence embedding derivation. Nonetheless, given that STS tasks typically categorize text pairs by degrees of semantic similarity, and Sentence-BERT approaches these tasks from a classification standpoint, neglecting the progressive relationships between categories, there exists a clear opportunity for improvement. As an illustration, consider an STS task with five categories, labeled consecutively from 1 to 5. Traditional classification strategies would yield identical loss for a sample scored at 2, irrespective of its prediction as 3 or 4, an approach evidently suboptimal.

To rectify such deficiency, this paper proposes a novel framework that converts multi-class STS tasks into regression problems, thus effectively capturing the progressive relationships between categories. For a given dataset, we first map its original labels to evenly spaced numerical values, ensuring that samples with higher similarity scores are assigned correspondingly greater values. Then, we set the number of nodes in the output layer to one, thereby enabling the model to produce a continuous prediction. Finally, the model parameters are updated according to the difference between predicted and actual scores.

Distinct from standard regression tasks, the ground truth within our transformed multi-category STS tasks manifest as a series of discrete points along the numerical axis. Therefore, instead of requiring precise matches to the target values, the floating-point predictions just need to be sufficiently close to get correctly classified. To accommodate this characteristic, we introduce a zero-gradient buffer zone to widely utilized L1 Loss and MSE Loss, unveiling two innovative loss functions: Translated ReLU and Smooth K2 Loss.

Comprehensive evaluations across seven STS benchmarks substantiate that our regression framework surpasses traditional classification strategies in handling multi-category STS tasks. Additionally, we find that our approach can further refine the performance of contrastive learning pre-trained models by utilizing filtered STS-B and SICK-R training sets. These findings highlight the effectiveness of our method and underscore the importance of harnessing task-specific data, an aspect often neglected in contrastive learning paradigms.

The main contributions of this study are outlined as follows:
\begin{itemize}
\item Building upon the foundation of Sentence-BERT, we develop a regression framework adept at modeling the progressive relationships between categories in multi-class STS tasks. This not only enhances performance but also, due to regression's intrinsic properties, simplifies the prediction process for K-category problems to require only a single output node, significantly minimizing the model's output layer parameter count.

\item We propose two novel loss functions, Translated ReLU and Smooth K2 Loss, specifically tailored to address classification problems involving progressive relationships between categories.

\item Through empirical evidence, we demonstrate that our strategy can be combined with leading contrastive learning pre-trained models, leveraging fine-grained annotated data to further improve their performance. This offers a new perspective for current research in STS and sentence embeddings.
\end{itemize}

\section{Related Work}

In this section, we primarily review three types of STS solutions that are directly relevant to our work:

\textbf{Siamese Neural Network Architectures}: These approaches \citep{Sentence-BERT-EMNLP-2019, InferSent-EMNLP-2017, Aug-SBERT-NAACL-2021}, proposed relatively earlier in the field, have been widely applied across various domains owing to their effectiveness on annotated corpus. Although their performance on the seven STS benchmarks (STS 12-16, STS-B, SICK-R) is generally inferior to contemporary contrastive learning methods, this disparity largely stems from the absence of task-specific training data. Thus, models have the flexibility to opt for alternative sources, such as Wikipedia \citep{SimCSE-EMNLP-2021} or NLI datasets \citep{snli-2015-EMNLP, mnli-2018-NAACL}, which adapt readily to triplet format. Given our goal of tackling multi-category STS tasks, our model architecture remains rooted in the Siamese network. However, in contrast to preceding efforts, we introduce an innovative regression framework specifically designed to capture the progressive relationships between categories.

\textbf{Contrastive Learning Fine-Tuning Methods}: Contrastive learning is currently the mainstream paradigm for addressing STS tasks, with substantial research exploring its integration with the fine-tuning of PLMs \citep{PromptBERT-EMNLP-2022, CoT-BERT-ICANN-2024}. However, contrastive learning loss functions, epitomized by InfoNCE Loss \citep{InfoNCE-2018}, concentrate exclusively on binary semantic categorization and are unable to fully utilize fine-grained labeled texts. Additionally, the necessity for large batch sizes to ensure negative sample diversity and prevent model collapse imposes significant computational demands. These two limitations are inherently difficult to overcome within contrastive learning itself, yet they are precisely the strengths of Sentence-BERT-style dual-tower models. Therefore, a primary objective of this paper is to investigate whether the performance of contrastive learning models can be further enhanced by incorporating traditional Siamese neural network architectures.

\textbf{Contrastive Learning Pre-Trained Models}: With the growing importance of embeddings in retrieval-augmented generation \citep{RAG-Survey-2024} and other application scenarios, more companies and institutions are dedicating efforts to developing specialized text representation models. These approaches generally adopt multi-stage contrastive learning strategies for network pre-training \citep{E5-2022, GTE-2023, BGE-SIGIR-2024}. Additionally, compared to large-scale generative PLMs, lightweight discriminative models that capture bidirectional dependencies are often more preferred. In our experiments section, we employ two state-of-the-art contrastive learning pre-trained models, Jina Embeddings v2 \citep{Jina2-2023} and Nomic Embed \citep{NomicEmbed-2024}. Both are BERT-based encoder architectures with a parameter size of 137 million.

\section{Methodology}

This section presents our methodological framework, beginning with a detailed exposition of the network architecture and its operational workflow in subsection~\ref{sec:architecture}. Then, in subsections~\ref{sec:translated_relu} and \ref{sec:smooth_k2_loss}, we introduce the two novel loss functions proposed in this study.

\subsection{Network Architecture}
\label{sec:architecture}

\begin{figure}[htbp]
\centering
\includegraphics[width=\linewidth]{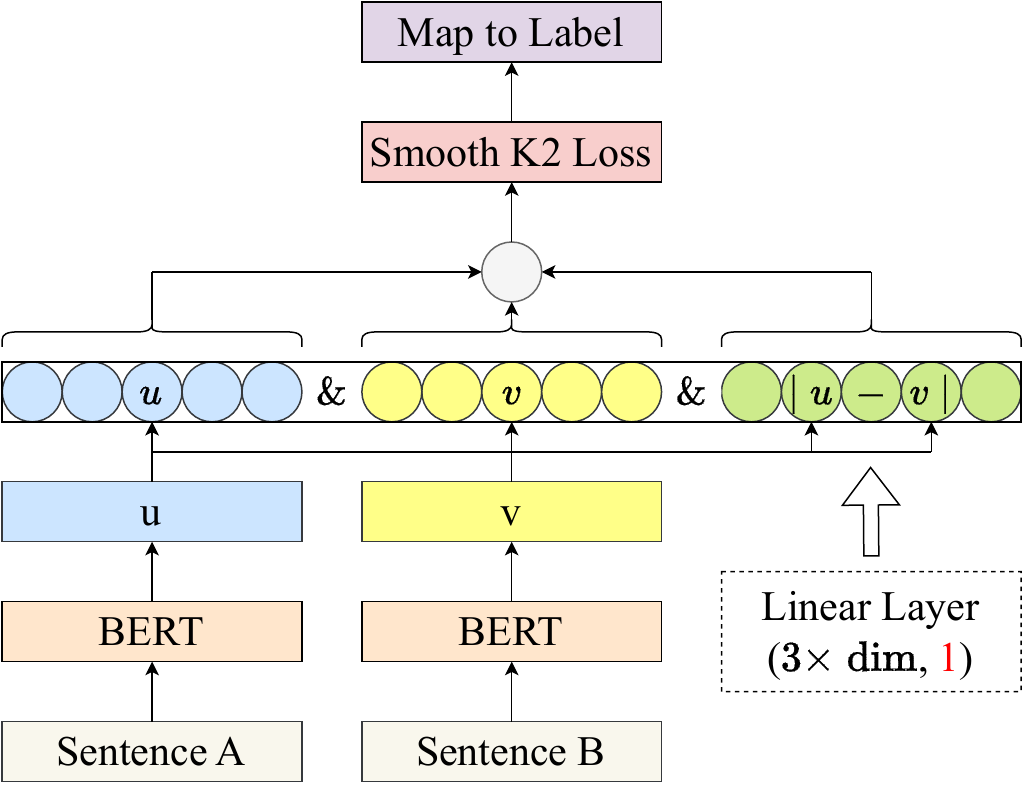}
\caption{Our Regression Framework. Here, the two BERT models share same parameters, with "dim" representing the embedding dimensions of $u$ and $v$.}
\label{fig:framework}
\end{figure}

As illustrated in Figure~\ref{fig:framework}, we utilize a Siamese neural network with shared parameters for encoding input sentences via BERT to obtain corresponding word embedding matrices. Subsequently, sentence embeddings, denoted as $u$ and $v$ for paired sentences $A$ and $B$, are derived through average pooling. These embeddings, both vectors of the hidden dimension, are then concatenated alongside their element-wise difference $|u-v|$ and passed through a fully connected layer with parameters sized at $3 \times \text{hidden\_dimension} \times 1$ to produce the model's predicted  similarity score.

Our method diverges from the original dual-tower structures employed by Sentence-BERT and InferSent \citep{InferSent-EMNLP-2017} in three critical aspects: 

1. We model STS tasks, characterized by a progressive relationship between categories, as regression problems. This is achieved by mapping labels from the original dataset to a sequence of incrementing numbers reflective of their similarity relations, thus conveying to the model that categories are not independent but progressively related.

2. Building on this, we streamline the output node count in the final fully connected layer to one, thereby enabling the model to directly yield a similarity score rather than a categorical probability distribution. Through this adjustment, for STS tasks with $K$ categories, we effectively reduce the parameter size of the output layer from $3 \times \text{hidden\_dimension} \times K$ to $3 \times \text{hidden\_dimension} \times 1$. In light of the expanding hidden layer dimensions in modern PLMs, this optimization can save considerable computational resources.

3. Unlike the classification-based approach of InferSent and Sentence-BERT, which assigns target classes for sentence pairs according to the highest probability, our regression framework categorizes based on the closeness between the predicted and actual values.

To better understand this process, consider an STS task with four categories: ``highly relevant,” ``moderately relevant,” ``slightly relevant,” and ``irrelevant.” After clarifying the progressive relationship between these categories, we would map them to four consecutive numbers 0, 1, 2, 3, respectively, ranging from ``irrelevant” to ``highly relevant.” This mapping strategy is flexible, allowing for task-specific adjustments in both numerical nodes and intervals. Furthermore, the mapped nodes do not necessarily have to be integers. Subsequently, we encode the paired sentences and compute their semantic similarity, resulting in a floating-point prediction. By rounding this value, it can be converted into the corresponding label. For instance, a prediction of 2.875 for a sample pair would be classified as ``highly relevant,” as it is closest to the boundary point of 3. Similarly, if a sample receives a predicted value of 1.333, it would be approximated to 1 and thus classified as ``slightly relevant," because 1.333 is closer to 1 among the four boundary points 0, 1, 2, 3.

Extending from the above examples, it can be seen that if the original labels are mapped to nodes spaced by $d$, as long as the difference between the model's prediction and the ground truth is less than $\frac{d}{2}$, the sample will be correctly classified. Specifically, for consecutive natural numbers, $d$ is equal to 1. However, conventional regression loss functions, represented by L1 Loss and MSE Loss, always enforce the model to exactly match the true value, a requirement that is unnecessary for our task scenario. Thus, we introduce a zero-gradient buffer zone into both functions, unveiling two new loss functions: Translated ReLU and Smooth K2 Loss.

\subsection{Translated ReLU}
\label{sec:translated_relu}

We first present Translated ReLU, mathematically formulated in Equation~\ref{eq:translated_relu}. Herein, $d$ represents the interval between mapped category labels. As previously discussed, when the difference between the model's predicted value and the ground truth is less than $\frac{d}{2}$, it signifies a correct classification of the sample. Traditional regression loss functions, however, mandate absolute congruence between predictions and true values, applying a penalty for any deviation. This stringent requirement to some extent diverts the model's focus from difficult samples that have not yet been correctly classified and ignores the inherent variability within classes. 
\begin{equation}
\label{eq:translated_relu}
\begin{aligned}
x \rightarrow & \text{abs}\text{(prediction - label)} \geq 0 \\
f(x) & = 
\begin{cases}
0 \quad x < x_0 \leq \frac{d}{2}
\\
k(x - x_0)\quad x_0\leq x
\end{cases} \\
f(x) & ={\max}\big{(}0,k(x-x_0)\big{)}
\end{aligned}
\end{equation}

To circumvent this limitation, we introduce an adjustable threshold hyperparameter $x_0$, and set the loss function to zero for values within $[0, x_0]$. This modification posits that a divergence less than $x_0$ between prediction and ground truth is deemed sufficiently precise, thus exempt from penalty or gradient update. For disparities exceeding $x_0$, Translated ReLU imposes a linear penalty. To maintain accurate classification, $x_0$ must not exceed $\frac{d}{2}$, with the interval between $x_0$ and $\frac{d}{2}$ acting as a margin akin to that in Hinge Loss. This margin can enhance model robustness by penalizing correctly predicted samples that lack adequate confidence. Additionally, a parameter $k$ is specified to control the slope of the function.

\begin{figure*}[htbp]
\centering
\includegraphics[width=\linewidth]{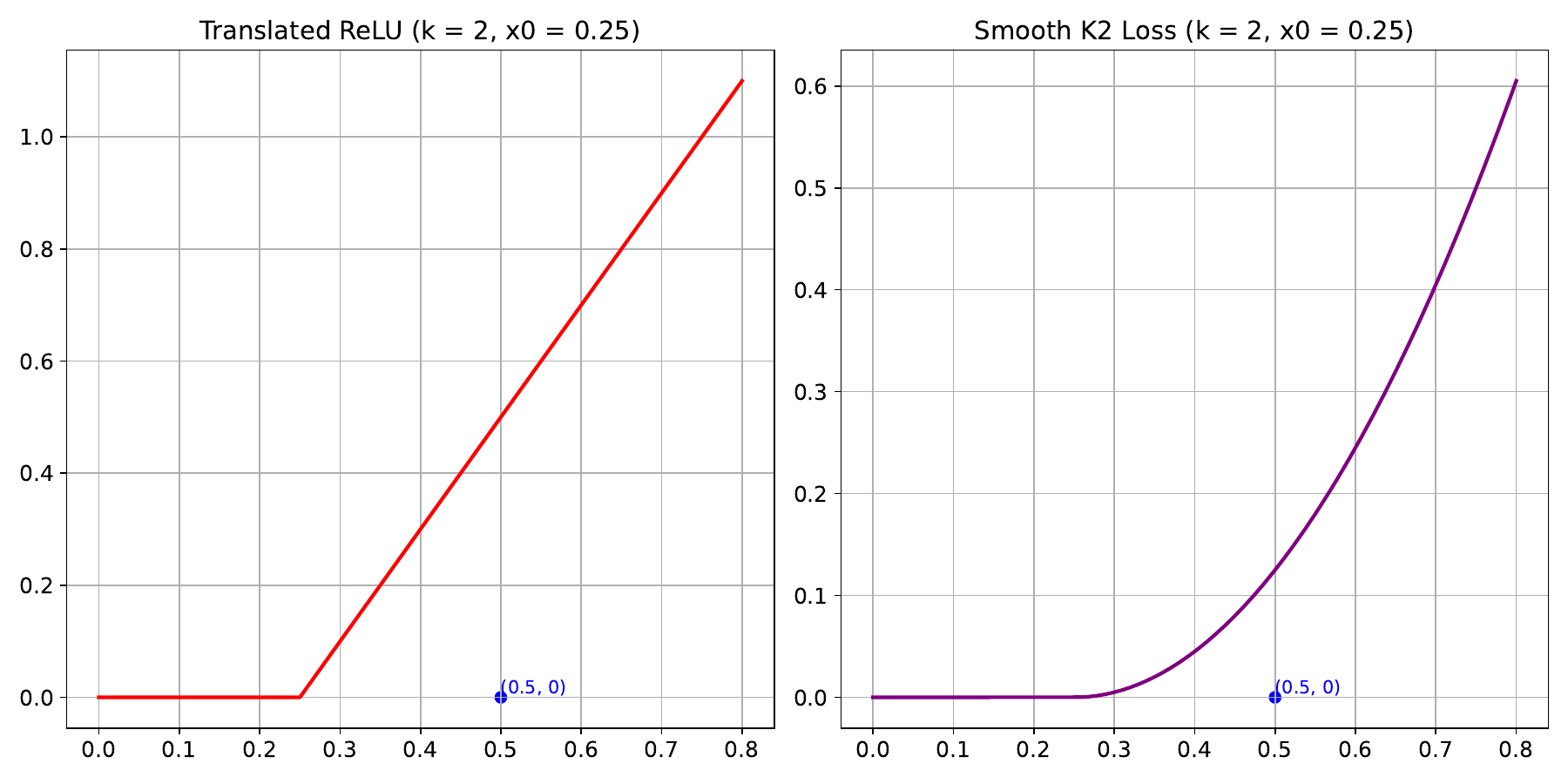}
\caption{Comparison of Translated ReLU and Smooth K2 Loss, both with $k = 2, x_0 = 0.25$.}
\label{fig:loss_functions}
\end{figure*}

The graphical depiction of Translated ReLU is exhibited on the left side of Figure~\ref{fig:loss_functions}, with parameters set to $k=2$ and $x_0=0.25$. This configuration resembles the ReLU activation function, albeit with a rightward translation. Our study employs Translated ReLU as a loss function and will compare its effects with those of L1 Loss in ensuing sections to demonstrate the significance of zero-gradient buffer zone for augmenting model performance.

\subsection{Smooth K2 Loss}
\label{sec:smooth_k2_loss}

Translated ReLU is characterized by its simplicity and efficacy. Nonetheless, we acknowledge its limitation pertaining to the abrupt lack of smoothness at the demarcation point $x = x_0$, alongside a constant gradient that fails to accommodate varying strengths of updates based on the distance between predictions and actual values. To address these concerns, we introduce another loss function termed Smooth K2 Loss to provide a smoother transition and a gradient that dynamically adjusts in accordance with the magnitude of discrepancy from the ground truth. The formulation and the derivative of Smooth K2 Loss are specified as follows:
\begin{equation}
\label{eq:smooth_k2_loss}
\begin{aligned}
x \rightarrow & \text{abs}\text{(prediction - label)} \geq 0 \\
f(x) & = 
\begin{cases}
0 \quad x < x_0 \leq \frac{d}{2}
\\
k(x^2-2x_0x+x_0^2)\quad x_0\leq x
\end{cases}
\\
\frac{\partial f(x)}{\partial x} &=
\begin{cases}
0 \quad x < x_0 \leq \frac{d}{2}
\\
2k(x-x_0)\quad x_0\leq x
\end{cases}
\end{aligned}
\end{equation}

Echoing the design of Translated ReLU, Smooth K2 Loss also incorporates a zero-gradient buffer zone, but exhibits a quadratic function for $x \geq x_0$, as illustrated on the right side of Figure~\ref{fig:loss_functions}. Given the differential mathematical underpinnings of these two loss functions, Smooth K2 Loss is recommended for scenarios with high-quality data and strong credibility. In contrast, when dealing with datasets that contain considerable noise, Translated ReLU may be a more suitable choice.

Additionally, prior to the application of Translated ReLU and Smooth K2 Loss, it is advisable to consider reassigning prediction values that transcend the defined category range to the nearest boundary. For instance, in a classification task where the category labels can be sequentially converted to 0, 1, 2 and 3, if the model predicts a value of 3.57 for a sample with an actual label of 3, this might be deemed acceptable and potentially obviate the need for a loss adjustment. This rationale stems from the observation that, despite the prediction's deviation exceeding $\frac{d}{2} = 0.5$, the absence of subsequent boundary points beyond 3 warrants a relaxation of this criterion.

\section{Experiment}

This section provides empirical validation of our regression framework and two innovative loss functions. We commence by comparing the performance of different modeling strategies for multi-category STS tasks and various loss functions (subsection~\ref{sec:exp_siamese}). Next, we demonstrate that, when supplemented with fine-grained training data, our Siamese neural network can effectively enhance the performance of contrastive learning PLMs (subsection~\ref{sec:exp_contrast}). Following this, we highlight the computational efficiency of our methodology (subsection~\ref{sec:memory}) and explore the influence of varying hyperparameter settings on model performance (subsection~\ref{sec:hyper}). Finally, subsection~\ref{sec:ablation} presents ablation studies on our network architecture.

\subsection{STS Performance Based on Traditional Discriminative Pre-Trained Models}
\label{sec:exp_siamese}
\begin{table*}[ht]
\centering
\resizebox{1.0\linewidth}{!}{
    \begin{tabular}{ccccccccc}
    \toprule
    \bf{Models} & \bf{STS-12} & \bf{STS-13} & \bf{STS-14} & \bf{STS-15} & \bf{STS-16} & \bf{STS-B} & \bf{SICK-R} & \bf{Avg.} \\      
    \midrule
    \multicolumn{9}{c}{\emph{Implementation on BERT$_{\text{base}}$}} \\
    {Sentence-BERT$_{\text{base}}$ $\clubsuit$} & 70.97 & 76.53 & 73.19 & 79.09 & 74.30 & 77.03 & 72.91 & 74.89 \\
    {BERT$_{\text{base}}$ $+$ Cross Entropy} & 70.01 & 71.18 & 70.10 & 78.37 & 72.92 & 74.88 & \bf{73.58} & 73.01 \\
    \midrule
    {BERT$_{\text{base}}$ $+$ L1 Loss} & 69.76 & 69.56 & 68.13 & 76.33 & 70.96 & 73.61 & 70.28 & 71.23 \\
    {BERT$_{\text{base}}$ $+$ Translated ReLU} & \bf{72.51} & 75.46 & 72.34 & 78.46 & 72.64 & 76.54 & 72.02 & 74.28 \\
    \midrule
    {BERT$_{\text{base}}$ $+$ MSE Loss} & 72.38 & 76.47 & 74.35 & 78.71 & 72.95 & 77.91 & 70.67 & 74.78 \\
    {BERT$_{\text{base}}$ $+$ Smooth K2 Loss} & 72.39 & \bf{78.33} & \bf{75.28} & \bf{80.26} & \bf{74.52} & \bf{78.78} & 72.65 & \bf{76.03} \\
    \midrule \midrule
    \multicolumn{9}{c}{\emph{Implementation on RoBERTa$_{\text{base}}$}} \\
    {Sentence-RoBERTa$_{\text{base}}$ $\clubsuit$} & 71.54 & 72.49 & 70.80 & 78.74 & 73.69 & 77.77 & \bf{74.46} & 74.21 \\
    {RoBERTa$_{\text{base}}$ $+$ Cross Entropy} & 71.15 & 74.29 & 72.66 & 79.44 & 74.12 &    76.56 & 73.02 & 74.46 \\ 
    \midrule
    {RoBERTa$_{\text{base}}$ $+$ L1 Loss} & 68.12 & 62.27 & 64.20 & 72.80 & 67.28 & 72.44 & 66.82 & 67.70 \\ 
    {RoBERTa$_{\text{base}}$ $+$ Translated ReLU} & 71.13 & 76.07 & 72.18 & 78.13 & 73.94 & 77.59 & 70.94 & 74.28 \\ 
    \midrule
    {RoBERTa$_{\text{base}}$ $+$ MSE Loss} & \bf{72.67} & 77.09 & 72.93 & 79.52 & 74.12 & \bf{77.88} & 69.85 & 74.87 \\
    {RoBERTa$_{\text{base}}$ $+$ Smooth K2 Loss} & 72.53 & \bf{78.28} & \bf{73.88} & \bf{80.88} & \bf{75.35} & 77.44  & 73.94 & \bf{76.04} \\    
    \bottomrule
    \end{tabular}%
}
\caption{Spearman's correlation scores for different methods across seven STS tasks. This table is partitioned to facilitate \textbf{a single variable comparison}. $\clubsuit$: results from \citep{Sentence-BERT-EMNLP-2019}.}
\label{tab:train_results}
\end{table*}

Our experimental setup here closely mirrors that of Sentence-BERT, leveraging fine-tuning on BERT or RoBERTa with a composite corpus derived from the SNLI and MNLI datasets. These NLI datasets categorize sentence pairs into three distinct classes: contradiction, neutral, and entailment. Sentence-BERT maps these classes to 0, 2, and 1, respectively, and employs a classification strategy for training \citep{Sentence-BERT-EMNLP-2019}. In contrast, our method sequentially maps contradiction, neutral, and entailment to 0, 1 and 2. This mapping reflects the natural order of semantic similarity, from least to most similar, thereby enabling our regression framework to better capture the progressive relationships between categories.
\begin{table}[htbp]
\renewcommand
\arraystretch{1.2}
\centering
\setlength{\tabcolsep}{2pt}
\begin{tabular}{cccc}
\hline
PLM & Loss & $k$ & $x_0$ \\
\hline
BERT$_{\text{base}}$ & Translated ReLU & 2.5 & 0.25\\
BERT$_{\text{base}}$ & Smooth K2 Loss & 2 & 0.25\\
RoBERTa$_{\text{base}}$ & Translated ReLU & 1 & 0.25\\
RoBERTa$_{\text{base}}$ & Smooth K2 Loss & 3 & 0.25\\
\hline
\end{tabular}
\caption{Hyperparameter configurations for our two loss functions when fine-tuning BERT and RoBERTa on the NLI dataset.}
\label{tab:k_x0}
\end{table}

For computational efficiency, we uniformly set the batch size to 16 and limit training to a single epoch, with model checkpoints saved based on performance metrics on the STS-B development set. The specific hyperparameter settings for Translated ReLU and Smooth K2 Loss are detailed in Table~\ref{tab:k_x0}. During evaluation, we assess the model's average Spearman correlation across seven STS tasks via the SentEval toolkit \citep{SentEval-LREC-2018}. The results of these experiments are summarized in Table~\ref{tab:train_results}, from which we distill insights along three pivotal aspects: 

1. \textbf{Classification Strategy vs. Regression Strategy:} Our regression framework, particularly when utilizing Smooth K2 Loss, yields an average Spearman correlation of 76.03 for BERT$_{\text{base}}$ and 76.04 for RoBERTa$_{\text{base}}$. These figures significantly outstrip those attained through Sentence-BERT and the classification strategy with Cross-Entropy Loss, highlighting the regression-based modeling's superiority in both reducing the output layer's parameter size and enhancing semantic discrimination in multi-category STS tasks.

2. \textbf{Efficacy of the Zero-Gradient Buffer Zone:} The adoption of Translated ReLU improves performance for both BERT and RoBERTa beyond what is achieved with L1 Loss. Similarly, employing Smooth K2 Loss surpasses MSE Loss on both PLMs. These comparisons underline the benefit of incorporating a zero-gradient buffer zone, which helps balance the model's attention across diverse samples in regression-modeled multi-category classification tasks.
\begin{table*}[ht]
\centering
\resizebox{1.0\linewidth}{!}{
    \begin{tabular}{ccccccccc}
    \toprule
    \bf{Models} & \bf{STS-12} & \bf{STS-13} & \bf{STS-14} & \bf{STS-15} & \bf{STS-16} & \bf{STS-B} & \bf{SICK-R} & \bf{Avg.} \\      
    \midrule
    {CT-SBERT$_{\text{base}}$ $\spadesuit$} & 74.84 & 83.20 & 78.07 & 83.84 & 77.93 & 81.46 & 76.42 & 79.39 \\
    {SimCSE-BERT$_{\text{base}}$ $\spadesuit$} & 75.30 & 84.67 & 80.19 & 85.40 & 80.82 & 84.25 & 80.39 & 81.57 \\
    {PromptBERT$_{\text{base}}$ $\heartsuit$} & 75.48 & 85.59 & 80.57 & 85.99 & 81.08 & 84.56 & 80.52 & 81.97 \\
    {PromCSE-BERT$_{\text{base}}$ $\diamondsuit$} & 75.58 & 84.33 & 79.67 & 85.79 & 81.24 & 84.25 & 80.79 & 81.81 \\ 
    \midrule
    {Nomic Embed Text v1} & 73.75 & 85.03 & 80.52 & 87.40 & 83.55 & 83.90 & 76.52 & 81.52 \\
    {Nomic Embed Text v1 + Contrast} & 76.10 & 85.79 & 80.58 & 87.35 & 83.54 & 85.16 &  72.33 & 81.55 \\ 
    {Nomic Embed Text v1 + Ours} & 73.06 & 86.63 & 81.06 & 87.67 & 83.43 & 85.18 & 82.75 & \bf 82.83 \\
    \midrule
    {Jina Embeddings v2} & 74.28 & 84.18 & 78.81 & 87.55 & 85.35 & 84.85 & 78.98 & 82.00 \\
    {Jina Embeddings v2 + Contrast} & 76.04 & 86.37 & 80.16 & 86.53 & 85.24 & 84.31 & 74.18 & 81.83 \\ 
    {Jina Embeddings v2 + Ours} & 75.17 & 86.10 & 79.96 & 88.44 & 85.01 & 86.83 &      83.34 & \bf 83.55 \\ 
    \bottomrule
    \end{tabular}%
}
\caption{Spearman's correlation coefficients of different methods across seven STS tasks. The "+Contrast" notation in the first column refers models further fine-tuned with contrastive learning. $\spadesuit$: results from \citep{SimCSE-EMNLP-2021}. $\heartsuit$: results from \citep{PromptBERT-EMNLP-2022}. $\diamondsuit$: results from \citep{PromCSE-EMNLP-2022}.}
\label{tab:tune_results}
\end{table*}

3. \textbf{Adaptive Gradients Aligned with Prediction Errors:} Models trained with Smooth K2 Loss outperform those utilizing Translated ReLU, and models employing MSE Loss exceed those with L1 Loss. This evidences the advantages of dispensing differentiated gradients in line with prediction-ground truth deviations, especially when leveraging high-quality datasets like NLI.

Collectively, these findings substantiate the merit of (1) adopting a regression framework for multi-class STS tasks and (2) enhancing traditional regression loss functions with a zero-gradient buffer zone to optimize model performance.

\subsection{STS Performance Based on Contrastive Learning Pre-Trained Models}
\label{sec:exp_contrast}

While the Siamese neural network, augmented by our regression framework and innovative loss functions, has exhibited significant performance improvements, a gap remains when compared to leading contrastive learning methods. To address this, we exploit the strengths of Siamese architectures in fully utilizing annotated data and explore whether it can be combined with top-performing contrastive learning models.

Jina Embeddings v2 \citep{Jina2-2023} and Nomic Embed \citep{NomicEmbed-2024} are two recently released embedding models that employ multi-stage contrastive learning strategies during pre-training, combining supervised and unsupervised approaches to optimize the networks. Both have achieved state-of-the-art results on the MTEB leaderboard \citep{MTEB-EACL-2023}. Therefore, if our method can further enhance the performance of these models, it would provide valuable insights for future research.

Among the seven STS benchmarks (STS12-16, STS-B, and SICK-R), STS-B and SICK-R come with their own training datasets. Specifically, STS-B contains sentence pairs with similarity scores ranging from 0 to 5, while SICK-R includes pairs with scores from 1 to 5. To ensure accurate evaluation, we performed strict data filtering to remove any training text pairs that appeared in the test sets. Details of this filtering process are provided in Appendix~\ref{appendix:filter}. We then applied a linear transformation, $5 \times \frac{\text{label}(z) - 1}{4}$, to convert all SICK-R training labels to the range [0, 5] and merged them with the filtered STS-B training set. This procedure resulted in a fine-grained, task-specific corpus containing 5,398 sentence pairs.

\begin{figure}[htbp]
\centering
\includegraphics[width=\linewidth]{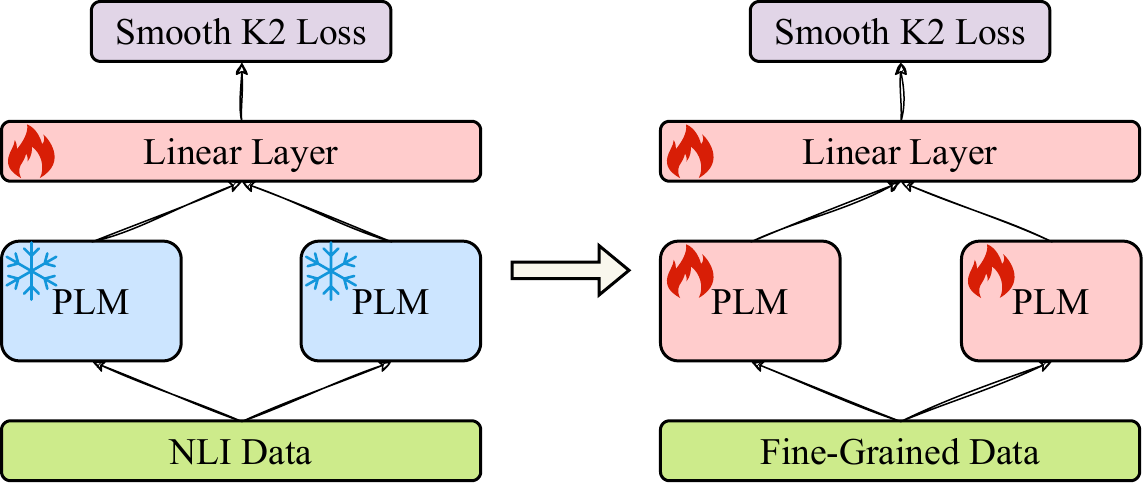}
\caption{Our two-stage fine-tuning process for contrastive learning pre-trained models. In the figure, modules highlighted in red are active during training and undergo backpropagation, while modules in blue are frozen and do not carry out updates.}
\label{fig:workflow}
\end{figure}

Since Jina Embeddings v2 and Nomic Embed have undergone pre-training on massive texts, their model parameters have favorable initial distributions. In contrast, our newly introduced linear layer is randomly initialized (Figure~\ref{fig:framework}). To facilitate effective joint training, we first freeze the entire PLM and only update the linear layer using the NLI dataset described in section~\ref{sec:exp_siamese}. After completing this step, we optimize both the PLM and the linear layer with the filtered STS training data. A schematic diagram of this workflow is shown in Figure~\ref{fig:workflow}. Throughout the entire procedure, Smooth K2 Loss is employed as the loss function.

The performance of Nomic Embed and Jina Embeddings v2 on the seven STS tasks before and after fine-tuning is presented in Table~\ref{tab:tune_results}. The results demonstrate that our network framework effectively enhances the performance of both models and surpasses BERT-based methods with comparable parameter sizes. Notably, we also test the impact of further updating the PLM using contrastive learning, which requires additional processing of the 5,398 training samples obtained earlier. To illustrate this, we take InfoNCE Loss \citep{InfoNCE-2018}, the most widely adopted contrastive learning loss function, as an example.

For any input sentence $x_i$, InfoNCE Loss computes the similarity between its encoding $f(x_i)$ and that of its positive instance $f(x_i^+)$ in the numerator, while aggregating the similarity calculations between $f(x_i)$ and other samples within the same batch in the denominator. This formulation aims to bring similar samples closer and push dissimilar ones apart. Equation~\ref{eq:info_nce} presents the standard expression of InfoNCE Loss, where $N$ represents the batch size and $\tau$ denotes a temperature hyperparameter.
\begin{equation}
\label{eq:info_nce}
     \mathcal{\ell}_{i} = - {\log} \frac{e^{{\cos}(f(x_i),f(x_i^+))/\tau}}{\sum_{j=1}^Ne^{{\cos}(f(x_i),f(x_j)^+)/\tau}}
\end{equation}

As indicated by Equation~\ref{eq:info_nce}, the only component of InfoNCE Loss that can be filled with labeled data is the similarity calculation between positive samples in the numerator. Consequently, contrastive learning is limited to utilizing only text pairs with the highest similarity ratings. To work within this constraint, we selected 1,543 samples from the 5,398 training pairs by adopting a threshold of 4.0 to filter out positive sample pairs. As it can be observed in Table~\ref{tab:tune_results}, after discarding such a large portion of annotation information, contrastive learning yields little improvement and may even lead to model collapse, causing performance degradation. In contexts where more detailed, domain-specific data is available, the shortcomings of contrastive learning in not being able to effectively harness multi-level label information, only performing coarse semantic distinctions, becomes more pronounced.

\subsection{Computational Resource Overhead}
\label{sec:memory}

In addition to its inability to fully leverage fine-grained annotated data, the high memory requirements of contrastive learning also pose a challenge for many researchers. In this section, we compare the computational resource consumption of our method with that of SimCSE during training, based on four 24GB NVIDIA GPUs. The results are summarized in Table~\ref{tab:memory}, where both BERT and RoBERTa are the base versions.

Despite setting the maximum sequence length for SimCSE to approximately 40\% of our method's default configuration, its memory usage remains significantly higher, reaching an astonishing 81GB. Thus, overall, our Siamese neural network strategy is more suitable for resource-constrained environments.
\begin{table}[htbp]
\centering
\begin{tabular}{c|c|c|c}
    \toprule
    \bf PLMs & \bf Method & Length & \bf Memory\\
    \midrule
    \multirow{2}{*}{BERT}
    & SimCSE & 100 & 81.30 GB\\
    & Ours & \bf 256 & \bf 41.27 GB\\
    \midrule
    \multirow{2}{*}{RoBERTa} 
    & SimCSE & 100 & 81.61 GB\\
    & Ours & \bf 256 & \bf 42.33 GB\\
    \bottomrule
\end{tabular}
\caption{Computational demands of our method compared to SimCSE during the training phase. The third column, "Length," represents the maximum sequence length supported by each model (cutoff length).}
\label{tab:memory}
\end{table}

\subsection{Impact of Different Hyperparameter Settings}
\label{sec:hyper}

In this study, we introduce two novel loss functions, Translated ReLU and Smooth K2 Loss, each characterized by two critical hyperparameters: $k$ and $x_0$. The parameter $k$ primarily controls the gradient of the loss function, while $x_0$ defines the tolerance threshold for model predictions. To discern the influence of these hyperparameters on model performance, we conducted a series of experiments across both traditional discriminative PLMs (BERT, RoBERTa) and the latest contrastive learning PLMs (Nomic Embed v1, Jina Embeddings v2).

The outcomes of these investigations are consolidated in Tables~\ref{tab:hyper}. Rather than executing an exhaustive grid search, initial values were selected based on our preliminary insights, followed by incremental adjustments. This implies that there may still be room for further improvement in our model's performance.
\begin{table}[htbp]
\centering
\resizebox{1.0\linewidth}{!}{
    \begin{tabular}{ccccc}
    \toprule
    \bf{PLM} & \bf{Loss} & \bf{$k$} & \bf{$x_0$} & \bf{Performance} \\      
    \midrule
    \multicolumn{5}{c}{\emph{Implementation on Traditional Discriminative PLMs}} \\
    {BERT$_{\text{base}}$} & Translated ReLU & 1.5 & 0.25 &  74.21 \\
    {BERT$_{\text{base}}$} & Translated ReLU & 2 & 0.25 &  74.21 \\
    {BERT$_{\text{base}}$} & Translated ReLU & 2.5 & 0.25 & \bf 74.28 \\
    \midrule
    {BERT$_{\text{base}}$} & Smooth K2 Loss & 3 & 0.25 & 75.75 \\
    {BERT$_{\text{base}}$} & Smooth K2 Loss & 2.5 & 0.25 & 75.89 \\
    {BERT$_{\text{base}}$} & Smooth K2 Loss & 2 & 0.25 & \bf 76.03 \\
    \midrule
    {RoBERTa$_{\text{base}}$} & Translated ReLU & 2 & 0.25 & 74.00 \\
    {RoBERTa$_{\text{base}}$} & Translated ReLU & 1.5 & 0.25 & 74.11 \\
    {RoBERTa$_{\text{base}}$} & Translated ReLU & 1 & 0.25 & \bf 74.28 \\
    \midrule
    {RoBERTa$_{\text{base}}$} & Smooth K2 Loss & 2.5 & 0.25 & 75.89 \\
    {RoBERTa$_{\text{base}}$} & Smooth K2 Loss & 3 & 0.2 & 75.90 \\
    {RoBERTa$_{\text{base}}$} & Smooth K2 Loss & 3 & 0.25 & \bf 76.04 \\
    \midrule \midrule
    \multicolumn{5}{c}{\emph{Implementation on Contrastive Learning PLMs}} \\
    {Nomic v1} & Smooth K2 Loss & 3.5 & 0.2 & 82.76 \\
    {Nomic v1} & Smooth K2 Loss & 2.5 & 0.2 & 82.79 \\
    {Nomic v1} & Smooth K2 Loss & 2 & 0.2 & 82.82 \\
    {Nomic v1} & Smooth K2 Loss & 3 & 0.2 & \bf 82.83 \\
    \midrule
    {Jina v2} & Smooth K2 Loss & 3 & 0.15 & 83.51 \\
    {Jina v2} & Smooth K2 Loss & 3 & 0.2 & 83.54 \\
    {Jina v2} & Smooth K2 Loss & 3.5 & 0.2 & \bf 83.55 \\
    {Jina v2} & Smooth K2 Loss & 4 & 0.2 & \bf 83.55 \\
    \bottomrule
    \end{tabular}
}
\caption{Average Spearman's correlation scores across seven STS tasks under different values of \(k\) and \(x_0\).}
\label{tab:hyper}
\end{table}

The experimental results from Table~\ref{tab:hyper} reveal minor fluctuations in model performance across diverse hyperparameter configurations, which affirms the resilience and robustness of our proposed methodology. This stability highlights the inherent adaptability of our regression framework as well as loss functions, suggesting their applicability to a wide range of modeling scenarios without necessitating extensive hyperparameter optimization.

\subsection{Ablation Studies}
\label{sec:ablation}

In section~\ref{sec:exp_siamese}, we initially demonstrated the effectiveness of our regression framework by comparing the performance of models utilizing both classification-based and regression-based strategies for multi-category STS tasks. Then, we elucidated the significance of zero-gradient buffer zones by evaluating the performance of models when selecting either Translated ReLU or L1 Loss, and Smooth K2 Loss or MSE Loss as the loss function. These comparisons directly align with the three core innovations of this paper and fulfill the role of ablation experiments.

Here, we extend our ablation study by evaluating our network architecture, as depicted in Figure~\ref{fig:framework}. Specifically, we seek to determine the necessity of concatenating $u$, $v$, and their element-wise difference $|u-v|$ in the final linear layer of the model. To this end, we employ both BERT and RoBERTa under the same experimental conditions outlined in section~\ref{sec:exp_siamese}, with the results presented in Table~\ref{tab:u-v}. The findings indicate that the concatenation method $(u, v, |u - v|)$ is the most effective for both PLMs, thus further validating the rationale behind our proposed scheme.
\begin{table}[htbp]
\renewcommand
\arraystretch{1.2}
\centering
\setlength{\tabcolsep}{2pt}
\begin{tabular}{ccc}
\hline
PLM & Concatenation & Spearman \\
\hline
BERT$_\text{base}$ & $(u, v)$ & 53.30 \\
BERT$_\text{base}$ & $(|u - v|)$ & 54.84 \\
BERT$_\text{base}$ & $(u, v, |u - v|)$ & \bf 76.03 \\
\hline
\hline
RoBERTa$_\text{base}$ & $(u, v)$ & 60.99 \\
RoBERTa$_\text{base}$ & $(|u - v|)$ & 59.10 \\
RoBERTa$_\text{base}$ & $(u, v, |u - v|)$ & \bf 76.04 \\
\hline
\end{tabular}
\caption{Average Spearman's correlation scores obtained by models on seven STS tasks with different concatenation methods in the final linear layer of our Siamese neural network architecture.}
\label{tab:u-v}
\end{table}

\section{Conclusion}

In this paper, we propose an innovative regression framework and develop two simple yet efficacious loss functions: Translated ReLU and Smooth K2 Loss, to address multi-class STS tasks. Compared to traditional classification approaches, our regression modeling strategy effectively captures the progressive relationships between categories, thereby achieving superior performance while reducing the the parameter count in the model's output layer.

Further empirical evidence demonstrates that our method can also be combined with leading contrastive learning models, leveraging fine-grained annotated data to further enhance their performance. Moreover, this approach proves to be more advantageous than continued fine-tuning through contrastive learning, both in terms of performance gains and computational efficiency. 

To support further research, we have made our code and model checkpoints publicly available.

\section*{Limitations}

Due to the lack of baselines and computational resource constraints, the experiments in this paper primarily focus on encoder-only discriminative models, rather than recently advanced generative pre-trained models (e.g. LLaMA \citep{LLaMA-2023}, Mistral \citep{Mistral-2023}). However, it is important to emphasize that, compared to mainstream generative PLMs, the models we selected—BERT, RoBERTa, Jina Embeddings v2, and Nomic Embed v1—have significantly fewer parameters. This results in higher inference efficiency, which is particularly advantageous in large-scale information retrieval and text clustering scenarios.

\bibliography{custom}

\begin{thebibliography}{31}
\providecommand{\natexlab}[1]{#1}

\bibitem[{Agirre et~al.(2015)Agirre, Banea, Cardie, Cer, Diab, Gonzalez-Agirre, Guo, Lopez-Gazpio, Maritxalar, Mihalcea, Rigau, Uria, and Wiebe}]{STS15}
Eneko Agirre, Carmen Banea, Claire Cardie, Daniel Cer, Mona Diab, Aitor Gonzalez-Agirre, Weiwei Guo, I{\~n}igo Lopez-Gazpio, Montse Maritxalar, Rada Mihalcea, German Rigau, Larraitz Uria, and Janyce Wiebe. 2015.
\newblock \href {https://doi.org/10.18653/v1/S15-2045} {{S}em{E}val-2015 task 2: Semantic textual similarity, {E}nglish, {S}panish and pilot on interpretability}.
\newblock In \emph{Proceedings of the 9th International Workshop on Semantic Evaluation ({S}em{E}val 2015)}, pages 252--263.

\bibitem[{Agirre et~al.(2014)Agirre, Banea, Cardie, Cer, Diab, Gonzalez-Agirre, Guo, Mihalcea, Rigau, and Wiebe}]{STS14}
Eneko Agirre, Carmen Banea, Claire Cardie, Daniel Cer, Mona Diab, Aitor Gonzalez-Agirre, Weiwei Guo, Rada Mihalcea, German Rigau, and Janyce Wiebe. 2014.
\newblock \href {https://doi.org/10.3115/v1/S14-2010} {{S}em{E}val-2014 task 10: Multilingual semantic textual similarity}.
\newblock In \emph{Proceedings of the 8th International Workshop on Semantic Evaluation ({S}em{E}val 2014)}, pages 81--91.

\bibitem[{Agirre et~al.(2016)Agirre, Banea, Cer, Diab, Gonzalez-Agirre, Mihalcea, Rigau, and Wiebe}]{STS16}
Eneko Agirre, Carmen Banea, Daniel Cer, Mona Diab, Aitor Gonzalez-Agirre, Rada Mihalcea, German Rigau, and Janyce Wiebe. 2016.
\newblock \href {https://doi.org/10.18653/v1/S16-1081} {{S}em{E}val-2016 task 1: Semantic textual similarity, monolingual and cross-lingual evaluation}.
\newblock In \emph{Proceedings of the 10th International Workshop on Semantic Evaluation ({S}em{E}val-2016)}, pages 497--511.

\bibitem[{Agirre et~al.(2012)Agirre, Cer, Diab, and Gonzalez-Agirre}]{STS12}
Eneko Agirre, Daniel Cer, Mona Diab, and Aitor Gonzalez-Agirre. 2012.
\newblock \href {https://aclanthology.org/S12-1051} {{S}em{E}val-2012 task 6: A pilot on semantic textual similarity}.
\newblock In \emph{*{SEM} 2012: The First Joint Conference on Lexical and Computational Semantics {--} Volume 1: Proceedings of the main conference and the shared task, and Volume 2: Proceedings of the Sixth International Workshop on Semantic Evaluation ({S}em{E}val 2012)}, pages 385--393.

\bibitem[{Agirre et~al.(2013)Agirre, Cer, Diab, Gonzalez-Agirre, and Guo}]{STS13}
Eneko Agirre, Daniel Cer, Mona Diab, Aitor Gonzalez-Agirre, and Weiwei Guo. 2013.
\newblock \href {https://aclanthology.org/S13-1004} {*{SEM} 2013 shared task: Semantic textual similarity}.
\newblock In \emph{Second Joint Conference on Lexical and Computational Semantics (*{SEM}), Volume 1: Proceedings of the Main Conference and the Shared Task: Semantic Textual Similarity}, pages 32--43.

\bibitem[{Bowman et~al.(2015)Bowman, Angeli, Potts, and Manning}]{snli-2015-EMNLP}
Samuel~R. Bowman, Gabor Angeli, Christopher Potts, and Christopher~D. Manning. 2015.
\newblock \href {https://doi.org/10.18653/v1/D15-1075} {A large annotated corpus for learning natural language inference}.
\newblock In \emph{Proceedings of the 2015 Conference on Empirical Methods in Natural Language Processing}, pages 632--642.

\bibitem[{Cer et~al.(2017)Cer, Diab, Agirre, Lopez-Gazpio, and Specia}]{STS-B}
Daniel Cer, Mona Diab, Eneko Agirre, I{\~n}igo Lopez-Gazpio, and Lucia Specia. 2017.
\newblock \href {https://doi.org/10.18653/v1/S17-2001} {{S}em{E}val-2017 task 1: Semantic textual similarity multilingual and crosslingual focused evaluation}.
\newblock In \emph{Proceedings of the 11th International Workshop on Semantic Evaluation ({S}em{E}val-2017)}, pages 1--14.

\bibitem[{Conneau and Kiela(2018)}]{SentEval-LREC-2018}
Alexis Conneau and Douwe Kiela. 2018.
\newblock \href {https://aclanthology.org/L18-1269} {{S}ent{E}val: An evaluation toolkit for universal sentence representations}.
\newblock In \emph{Proceedings of the Eleventh International Conference on Language Resources and Evaluation ({LREC} 2018)}.

\bibitem[{Conneau et~al.(2017)Conneau, Kiela, Schwenk, Barrault, and Bordes}]{InferSent-EMNLP-2017}
Alexis Conneau, Douwe Kiela, Holger Schwenk, Lo{\"\i}c Barrault, and Antoine Bordes. 2017.
\newblock \href {https://doi.org/10.18653/v1/D17-1070} {Supervised learning of universal sentence representations from natural language inference data}.
\newblock In \emph{Proceedings of the 2017 Conference on Empirical Methods in Natural Language Processing}, pages 670--680.

\bibitem[{Devlin et~al.(2019)Devlin, Chang, Lee, and Toutanova}]{BERT-NAACL-2019}
Jacob Devlin, Ming-Wei Chang, Kenton Lee, and Kristina Toutanova. 2019.
\newblock \href {https://doi.org/10.18653/v1/N19-1423} {{BERT}: Pre-training of deep bidirectional transformers for language understanding}.
\newblock In \emph{Proceedings of the 2019 Conference of the North {A}merican Chapter of the Association for Computational Linguistics: Human Language Technologies, Volume 1 (Long and Short Papers)}, pages 4171--4186.

\bibitem[{Gao et~al.(2021)Gao, Yao, and Chen}]{SimCSE-EMNLP-2021}
Tianyu Gao, Xingcheng Yao, and Danqi Chen. 2021.
\newblock \href {https://doi.org/10.18653/v1/2021.emnlp-main.552} {{S}im{CSE}: Simple contrastive learning of sentence embeddings}.
\newblock In \emph{Proceedings of the 2021 Conference on Empirical Methods in Natural Language Processing}, pages 6894--6910.

\bibitem[{G{\"u}nther et~al.(2023)G{\"u}nther, Ong, Mohr, Abdessalem, Abel, Akram, Guzman, Mastrapas, Sturua, Wang et~al.}]{Jina2-2023}
Michael G{\"u}nther, Jackmin Ong, Isabelle Mohr, Alaeddine Abdessalem, Tanguy Abel, Mohammad~Kalim Akram, Susana Guzman, Georgios Mastrapas, Saba Sturua, Bo~Wang, et~al. 2023.
\newblock Jina embeddings 2: 8192-token general-purpose text embeddings for long documents.
\newblock \emph{arXiv preprint arXiv:2310.19923}.

\bibitem[{Jiang et~al.(2023)Jiang, Sablayrolles, Mensch, Bamford, Chaplot, Casas, Bressand, Lengyel, Lample, Saulnier et~al.}]{Mistral-2023}
Albert~Q Jiang, Alexandre Sablayrolles, Arthur Mensch, Chris Bamford, Devendra~Singh Chaplot, Diego de~las Casas, Florian Bressand, Gianna Lengyel, Guillaume Lample, Lucile Saulnier, et~al. 2023.
\newblock Mistral 7b.
\newblock \emph{arXiv preprint arXiv:2310.06825}.

\bibitem[{Jiang et~al.(2022{\natexlab{a}})Jiang, Jiao, Huang, Zhang, Wang, Zhuang, Wei, Huang, Deng, and Zhang}]{PromptBERT-EMNLP-2022}
Ting Jiang, Jian Jiao, Shaohan Huang, Zihan Zhang, Deqing Wang, Fuzhen Zhuang, Furu Wei, Haizhen Huang, Denvy Deng, and Qi~Zhang. 2022{\natexlab{a}}.
\newblock \href {https://doi.org/10.18653/v1/2022.emnlp-main.603} {{P}rompt{BERT}: Improving {BERT} sentence embeddings with prompts}.
\newblock In \emph{Proceedings of the 2022 Conference on Empirical Methods in Natural Language Processing}, pages 8826--8837.

\bibitem[{Jiang et~al.(2022{\natexlab{b}})Jiang, Zhang, and Wang}]{PromCSE-EMNLP-2022}
Yuxin Jiang, Linhan Zhang, and Wei Wang. 2022{\natexlab{b}}.
\newblock \href {https://doi.org/10.18653/v1/2022.findings-emnlp.220} {Improved universal sentence embeddings with prompt-based contrastive learning and energy-based learning}.
\newblock In \emph{Findings of the Association for Computational Linguistics: EMNLP 2022}, pages 3021--3035. Association for Computational Linguistics.

\bibitem[{Li et~al.(2023)Li, Zhang, Zhang, Long, Xie, and Zhang}]{GTE-2023}
Zehan Li, Xin Zhang, Yanzhao Zhang, Dingkun Long, Pengjun Xie, and Meishan Zhang. 2023.
\newblock Towards general text embeddings with multi-stage contrastive learning.
\newblock \emph{arXiv preprint arXiv:2308.03281}.

\bibitem[{Liu et~al.(2023)Liu, Liu, Wang, Wang, Wu, Xian, Zhao, Chen, and Yan}]{RankCSE-ACL-2023}
Jiduan Liu, Jiahao Liu, Qifan Wang, Jingang Wang, Wei Wu, Yunsen Xian, Dongyan Zhao, Kai Chen, and Rui Yan. 2023.
\newblock \href {https://doi.org/10.18653/v1/2023.acl-long.771} {{R}ank{CSE}: Unsupervised sentence representations learning via learning to rank}.
\newblock In \emph{Proceedings of the 61st Annual Meeting of the Association for Computational Linguistics (Volume 1: Long Papers)}, pages 13785--13802.

\bibitem[{Liu et~al.(2019)Liu, Ott, Goyal, Du, Joshi, Chen, Levy, Lewis, Zettlemoyer, and Stoyanov}]{RoBERTa-2019}
Yinhan Liu, Myle Ott, Naman Goyal, Jingfei Du, Mandar Joshi, Danqi Chen, Omer Levy, Mike Lewis, Luke Zettlemoyer, and Veselin Stoyanov. 2019.
\newblock Roberta: A robustly optimized bert pretraining approach.
\newblock \emph{arXiv preprint arXiv:1907.11692}.

\bibitem[{Marelli et~al.(2014)Marelli, Menini, Baroni, Bentivogli, Bernardi, and Zamparelli}]{SICK-R}
Marco Marelli, Stefano Menini, Marco Baroni, Luisa Bentivogli, Raffaella Bernardi, and Roberto Zamparelli. 2014.
\newblock \href {http://www.lrec-conf.org/proceedings/lrec2014/pdf/363_Paper.pdf} {A {SICK} cure for the evaluation of compositional distributional semantic models}.
\newblock In \emph{Proceedings of the Ninth International Conference on Language Resources and Evaluation ({LREC}'14)}, pages 216--223.

\bibitem[{Muennighoff et~al.(2023)Muennighoff, Tazi, Magne, and Reimers}]{MTEB-EACL-2023}
Niklas Muennighoff, Nouamane Tazi, Loic Magne, and Nils Reimers. 2023.
\newblock \href {https://doi.org/10.18653/v1/2023.eacl-main.148} {{MTEB}: Massive text embedding benchmark}.
\newblock In \emph{Proceedings of the 17th Conference of the European Chapter of the Association for Computational Linguistics}, pages 2014--2037. Association for Computational Linguistics.

\bibitem[{Nussbaum et~al.(2024)Nussbaum, Morris, Duderstadt, and Mulyar}]{NomicEmbed-2024}
Zach Nussbaum, John~X Morris, Brandon Duderstadt, and Andriy Mulyar. 2024.
\newblock Nomic embed: Training a reproducible long context text embedder.
\newblock \emph{arXiv preprint arXiv:2402.01613}.

\bibitem[{Oord et~al.(2018)Oord, Li, and Vinyals}]{InfoNCE-2018}
Aaron van~den Oord, Yazhe Li, and Oriol Vinyals. 2018.
\newblock Representation learning with contrastive predictive coding.
\newblock \emph{arXiv preprint arXiv:1807.03748}.

\bibitem[{Reimers and Gurevych(2019)}]{Sentence-BERT-EMNLP-2019}
Nils Reimers and Iryna Gurevych. 2019.
\newblock \href {https://doi.org/10.18653/v1/D19-1410} {Sentence-{BERT}: Sentence embeddings using {S}iamese {BERT}-networks}.
\newblock In \emph{Proceedings of the 2019 Conference on Empirical Methods in Natural Language Processing and the 9th International Joint Conference on Natural Language Processing (EMNLP-IJCNLP)}, pages 3982--3992.

\bibitem[{Thakur et~al.(2021)Thakur, Reimers, Daxenberger, and Gurevych}]{Aug-SBERT-NAACL-2021}
Nandan Thakur, Nils Reimers, Johannes Daxenberger, and Iryna Gurevych. 2021.
\newblock \href {https://doi.org/10.18653/v1/2021.naacl-main.28} {Augmented {SBERT}: Data augmentation method for improving bi-encoders for pairwise sentence scoring tasks}.
\newblock In \emph{Proceedings of the 2021 Conference of the North American Chapter of the Association for Computational Linguistics: Human Language Technologies}, pages 296--310.

\bibitem[{Touvron et~al.(2023)Touvron, Lavril, Izacard, Martinet, Lachaux, Lacroix, Rozi{\`e}re, Goyal, Hambro, Azhar et~al.}]{LLaMA-2023}
Hugo Touvron, Thibaut Lavril, Gautier Izacard, Xavier Martinet, Marie-Anne Lachaux, Timoth{\'e}e Lacroix, Baptiste Rozi{\`e}re, Naman Goyal, Eric Hambro, Faisal Azhar, et~al. 2023.
\newblock Llama: Open and efficient foundation language models.
\newblock \emph{arXiv preprint arXiv:2302.13971}.

\bibitem[{Wang et~al.(2022)Wang, Yang, Huang, Jiao, Yang, Jiang, Majumder, and Wei}]{E5-2022}
Liang Wang, Nan Yang, Xiaolong Huang, Binxing Jiao, Linjun Yang, Daxin Jiang, Rangan Majumder, and Furu Wei. 2022.
\newblock Text embeddings by weakly-supervised contrastive pre-training.
\newblock \emph{arXiv preprint arXiv:2212.03533}.

\bibitem[{Williams et~al.(2018)Williams, Nangia, and Bowman}]{mnli-2018-NAACL}
Adina Williams, Nikita Nangia, and Samuel Bowman. 2018.
\newblock \href {https://doi.org/10.18653/v1/N18-1101} {A broad-coverage challenge corpus for sentence understanding through inference}.
\newblock In \emph{Proceedings of the 2018 Conference of the North {A}merican Chapter of the Association for Computational Linguistics: Human Language Technologies, Volume 1 (Long Papers)}, pages 1112--1122.

\bibitem[{Xiao et~al.(2024)Xiao, Liu, Zhang, Muennighoff, Lian, and Nie}]{BGE-SIGIR-2024}
Shitao Xiao, Zheng Liu, Peitian Zhang, Niklas Muennighoff, Defu Lian, and Jian-Yun Nie. 2024.
\newblock C-pack: Packed resources for general chinese embeddings.
\newblock In \emph{Proceedings of the 47th International ACM SIGIR Conference on Research and Development in Information Retrieval}, pages 641--649.

\bibitem[{Zhang et~al.(2024)Zhang, Chang, and Li}]{CoT-BERT-ICANN-2024}
Bowen Zhang, Kehua Chang, and Chunping Li. 2024.
\newblock Cot-bert: Enhancing unsupervised sentence representation through chain-of-thought.
\newblock In \emph{International Conference on Artificial Neural Networks}, pages 148--163. Springer.

\bibitem[{Zhang et~al.(2023)Zhang, Zhang, Li, and Smola}]{AutoCoT-ICLR-2023}
Zhuosheng Zhang, Aston Zhang, Mu~Li, and Alex Smola. 2023.
\newblock \href {https://openreview.net/pdf?id=5NTt8GFjUHkr} {Automatic chain of thought prompting in large language models}.
\newblock In \emph{The Eleventh International Conference on Learning Representations, {ICLR} 2023, Kigali, Rwanda, May 1-5, 2023}.

\bibitem[{Zhao et~al.(2024)Zhao, Zhang, Yu, Wang, Geng, Fu, Yang, Zhang, and Cui}]{RAG-Survey-2024}
Penghao Zhao, Hailin Zhang, Qinhan Yu, Zhengren Wang, Yunteng Geng, Fangcheng Fu, Ling Yang, Wentao Zhang, and Bin Cui. 2024.
\newblock Retrieval-augmented generation for ai-generated content: A survey.
\newblock \emph{arXiv preprint arXiv:2402.19473}.

\end{thebibliography}

\appendix

\section{Data Filtering Method}
\label{appendix:filter}

As mentioned in section~\ref{sec:exp_contrast}, before applying the STS-B and SICK-R training sets for model updates, we implemented strict data filtering to ensure that no sentence pairs present in the test sets would appear in the fine-tuning corpus.

To elaborate on this process, we first take the SICK-R dataset as an example to illustrate the standard format of STS datasets. As shown in Table~\ref{tab:sick_r}, each sample consists of two text strings, "sentence 1" and "sentence 2," along with a floating-point number "score" that indicates the semantic similarity between them. We denote these as "s1," "s2," and "r," respectively. 

Then, for any sentence pair \((s1_i, s2_i, r_i)\) within the STS-B or SICK-R training set, if a sample \((s1_j, s2_j, r_j)\) exists in the test sets of STS12-16, STS-B, or SICK-R such that \(s1_i = s1_j\) and \(s2_i = s2_j\), or \(s1_i = s2_j\) and \(s2_i = s1_j\), we treat them as duplicates and remove the corresponding sentence pair from the training data. It should be noted that the entire process is conducted without any modifications to the test sets.

This filtering mechanism is stringent, as we do not take into account whether \(r_i\) and \(r_j\) are equal. In other words, as long as a sentence pair appears in both the training and test sets, it is removed from the training corpus, regardless of whether the similarity scores are identical. Under this protocol, even within the SICK-R dataset itself, there are instances where samples from the training and test sets overlap. Examples in Table~\ref{tab:sick_r} illustrate such cases. The goal of this approach is to maximize the model's generalization ability.
\begin{table*}[htbp]
\centering
\begin{tabular}{>{\centering\arraybackslash}m{0.43\linewidth}|>{\centering\arraybackslash}m{0.43\linewidth}|>{\centering\arraybackslash}c}
\toprule
sentence 1 & sentence 2 & score \\
\toprule \toprule
\multicolumn{3}{c}{\emph{Sentence pairs in the SICK-R \textbf{training} set}} \\
\midrule \midrule
A man in a blue jumpsuit is courageously performing a wheelie on a motorcycle & 
The man is doing a wheelie with a motorcycle on ground which is mostly barren & 
4.1 \\
\midrule
The tan dog is watching a brown dog that is swimming in a pond &
A pet dog is standing on the bank and is looking at another brown dog in the pond &
4.3 \\
\midrule \midrule
\multicolumn{3}{c}{\emph{Sentence pairs in the SICK-R \textbf{test} set}} \\
\midrule \midrule
The man is doing a wheelie with a motorcycle on ground which is mostly barren & 
A man in a blue jumpsuit is courageously performing a wheelie on a motorcycle & 
3.7 \\
\midrule
A pet dog is standing on the bank and is looking at another brown dog in the pond &
The tan dog is watching a brown dog that is swimming in a pond &
3.6 \\
\bottomrule
\end{tabular}
\caption{Samples from the SICK-R training and test sets. In these samples, text pair duplication occurs. Thus, the corresponding training samples are removed from the fine-tuning corpus used in section~\ref{sec:exp_contrast}.}
\label{tab:sick_r}
\end{table*}

\end{document}